\documentclass[conference]{IEEEtran}
\IEEEoverridecommandlockouts
\usepackage{cite}
\usepackage{amsmath,amssymb,amsfonts}
\usepackage{textcomp}
\usepackage{xcolor}
\usepackage{algorithm}
\usepackage{algpseudocode}
\usepackage{amsmath}

\usepackage{algorithm}
\usepackage{algpseudocode}

\usepackage{graphics}
\usepackage{caption}
\usepackage{subcaption}
\usepackage{graphicx}
\usepackage{tabularx}

\def\BibTeX{{\rm B\kern-.05em{\sc i\kern-.025em b}\kern-.08em
    T\kern-.1667em\lower.7ex\hbox{E}\kern-.125emX}}
    
\usepackage[overlay,absolute]{textpos}

\begin{document}

\begin{textblock*}{10in}(25mm, 10mm)
{\textbf{Ref:} \emph{International Joint Conference on Neural Networks (IJCNN)}, pages 1--8, Budapest, Hungary, July 2019.}
\end{textblock*}

\title{Stealing Knowledge from Protected Deep Neural Networks Using Composite Unlabeled Data}

\author{\IEEEauthorblockN{Itay Mosafi}
\IEEEauthorblockA{\textit{Department of Computer Science} \\
\textit{Bar-Ilan University}\\
Ramat-Gan, Israel \\
itay.mosafi@gmail.com}
\and
\IEEEauthorblockN{Eli (Omid) David}
\IEEEauthorblockA{\textit{Department of Computer Science} \\
\textit{Bar-Ilan University}\\
Ramat-Gan, Israel \\
mail@elidavid.com}
\and
\IEEEauthorblockN{Nathan S. Netanyahu}
\IEEEauthorblockA{\textit{Department of Computer Science} \\
\textit{Bar-Ilan University}\\
Ramat-Gan, Israel \\
nathan@cs.biu.ac.il}}

\maketitle

\makeatletter
\def\thickhline{%
  \noalign{\ifnum0=`}\fi\hrule \@height \thickarrayrulewidth \futurelet
   \reserved@a\@xthickhline}
\def\@xthickhline{\ifx\reserved@a\thickhline
               \vskip\doublerulesep
               \vskip-\thickarrayrulewidth
             \fi
      \ifnum0=`{\fi}}
\makeatother

\newlength{\thickarrayrulewidth}
\setlength{\thickarrayrulewidth}{2\arrayrulewidth}

\newcommand\setrow[1]{\gdef\rowmac{#1}#1\ignorespaces}
\newcommand\clearrow{\global\let\rowmac\relax}
\clearrow

\begin{abstract}
As state-of-the-art deep neural networks are deployed at the core of more advanced AI-based products and services, the incentive for copying them (i.e., their intellectual properties) by rival adversaries is expected to increase considerably over time. The best way to extract or steal knowledge from such networks is by querying them using a large dataset of random samples and recording their output, followed by training a {\it student} network to mimic these outputs, without making any assumption about the original networks. The most effective way to protect against such a mimicking attack is to provide only the classification result, without confidence values associated with the softmax layer.

In this paper, we present a novel method for generating composite images for attacking a {\it mentor} neural network using a student model. Our method assumes no information regarding the mentor's training dataset, architecture, or weights. Further assuming no information regarding the mentor's softmax output values, our method successfully mimics the given neural network and steals all of its knowledge. We also demonstrate that our student network (which copies the mentor) is impervious to watermarking protection methods, and thus would not be detected as a stolen model. 

Our results imply, essentially, that all current neural networks are vulnerable to mimicking attacks, even if they do not divulge anything but the most basic required output, and that the student model which mimics them cannot be easily detected and singled out as a stolen copy using currently available techniques.

\end{abstract}

\section{Introduction}
In recent years deep neural networks (DNNs) have been used very effectively in a wide range of applications. Since these models have achieved remarkable results, redefining state-of-the-art solutions for various problems, they have become the ``go-to solution'' for many challenging real-world problems, e.g., object recognition~\cite{girshick2015fast,ren2015faster}, object segmentation~\cite{redmon2016you}, autonomous driving~\cite{chen2015deepdriving}, automatic text translation~\cite{LuongPM15}, cybersecurity~\cite{rosenberg2017}, etc. 

Training a state-of-the-art deep neural network requires designing the network architecture, collecting and preprocessing data, and accessing hardware resources, in particular graphics processing units (GPUs) capable of training such models. Additionally, training such networks requires a substantial amount of trial and error. For these reasons, such trained models are highly valuable, but at the same time they could be the target of attacks by adversaries (e.g., a competitor) who might try to duplicate the model and the entire sensitive intellectual property involved, without going through the tedious and expensive process of developing the models by themselves. By doing so, all the trouble of data collection, acquiring computing resources, and the valuable time required for training the models are spared by the attacker. As state-of-the-art DNNs are used more extensively in real-world products, the prevalence of such attacks is expected to increase over the next few years.

An attacker has two main options for acquiring a trained model: (1) Acquiring the raw model from the owner's private network, which would be a risky criminal offense that requires a complicated cyber attack on the owner's network; and (2) training a student model that mimics the original mentor model. That is, the attacker could query the original mentor using a dataset of samples, and train the student model to mimic the output of the mentor model for each of the samples. The second option assumes that the mentor is a black box, i.e., there is no knowledge of its architecture, no access to the training data used for training it, and no information regarding the trained model's weights. We only have access to the model's predictions (inference) for a given input. Thus, such a mentor would effectively teach a student how to mimic it, by providing its output for different inputs.

In order for mimicking to succeed, a key element is to utilize the certainty level of a model on a given input, i.e., its softmax distribution values. This knowledge is important for the training of the student network. For example, in case of a binary classification, classifying an image as category $i$ with 99\% confidence and as category $j$ with 1\% confidence is much more informative than classifying it to category $i$ with, say, 51\% confidence and to category $j$ with 49\% confidence. This knowledge is valuable and much more informative than the predicted category alone, which in both cases is $i$. This confidence value (obtained through the softmax output layer) also reveals how the model perceives this specific image, and to what extent the predictions for categories $i$ and $j$ are similar.

In order to protect against such a mimicking attack, a trained model may hide this confidence information, by simply returning only the index with the maximal confidence, without providing the actual confidence levels (i.e., the softmax values are concealed, while the output contains merely the predicted class). Although such a model would substantially limit the success of a student model using a standard mimicking attack, we provide in this paper a novel method, by querying the mentor with {\it composite} images, such that the student effectively elicits the mentor's knowledge, even if the mentor provides the predicted class only.

The rest of the paper is organized as follows. Section II reviews previous methods used for network distilling and mimicking. Section III describes our new approach for a successful mimicking attack on a mentor which does not provide softmax outputs. Section IV presents our experimental results. Finally, Section IV makes concluding remarks.

\section{Background}
Our composite method is engineered to steal a neural network's intellectual property. We focus in this section on stealing both a DNN's field domain and the DNN's protection methods used to prevent a model from being stolen.
Our method can bypass all available protection methods and steal a model while carrying no marks that can identify the created model as stolen. 

\subsection{Watermarking}
The idea of {\it watermarking} is well known. It was originally invented in order to protect digital media from being stolen. The idea relies on inserting a unique modification or signature not visible to the human eye. This allows to prove legitimate ownership by presenting that the owner's unique signature is embedded into the digital media \cite{lee2009advanced}, \cite{tian2003reversible}.

With the same goal in mind, embedding a unique signature into a model and subsequently identifying the stolen model based on that signature, some new techniques were invented.
A method to embed a signature into the model's weights is described in \cite{uchida2017embedding}; it allows for the identification of the unique signature by examining the model's weights. This method assumes that the model and its parameters are available for examination. Unfortunately, in most cases the model's weights are not publicly available; an individual could offer an API-based service which uses the stolen model, while still keeping the model's parameters hidden from the user. Therefore, this method is not sufficient.

Another method \cite{merrer2017adversarial} proposes a zero-bit watermarking algorithm that makes use of adversaries' examples. It enables the authentication of the model's ownership using a set of queries. The authors rely on predefined examples that give certain answers. By showing that these exact same answers are obtained using $N$ queries, one can authenticate their ownership over the model. However, this idea may be problematic, since these queries are not unique and there can be infinitely many of them. An individual can generate queries for which a model outputs certain answers that match the original queries. In doing so, anyone can claim ownership. Furthermore, it is possible that different adversaries will have a different set of queries which give the exact predefined answers.

Some more recent papers~\cite{nagai2018digital},~\cite{rouhanideepsigns} offer a methodology that allows for inserting a digital watermarking into a deep learning (DL) model without harming the performance and with high model pruning resistance. In~\cite{watermarking2018} a method of inserting watermarking into a model is presented. Specifically, it allows to identify a stolen model even if it is used via an {\it application programming interface} (API) and returns only the predicted label. It is done by defining a certain hidden ``key" which can be a certain shape or noise integrated into a part of the training set. When the model receives an input containing the key, it will predict with high certainty a completely unrelated label. Thus, it is possible to use some available APIs by sending them an image integrated with the hidden key. If the result is odd and the unrelated label is triggered, it may be an indication that this model is stolen. Our method is resistant to this protection mechanism, as its learning is based on the predictions of the mentor. Specifically, our training is based on random combinations of inputs, i.e., the chances of sending the mentor a hidden key that will trigger the unrelated label mechanism is negligible. We can train and gain the important knowledge of such a model without learning the watermarks, thereby assuring that our model would not be identified as stolen when provided a hidden key as input.

Finally,~\cite{hitaj2018have} shows that a malicious adversary, even in scenarios where the watermark is difficult to remove, can still evade the verification by the legitimate owners. In conclusion, even the most advanced watermarking methods are still not good enough to properly protect a neural network from being stolen. Our composite method overcomes all of the above defense mechanisms.

\subsection{Attack Mechanisms}
As previously explained, trained deep neural networks are extremely valuable and worth protecting. Naturally, a lot of research has been done on attacking such networks and stealing their knowledge. In~\cite{fredrikson2015model}, \cite{tramer2016stealing} an attack method exploiting the confidence level of a model is presented. The assumption that the confidence level is available is too lenient, as it can be easily blocked by returning merely the predicted label. Our composite method shows how to successfully steal a model which does not reveal its confidence level(s).

In~\cite{correia2018copycat} it is shown how to steal the knowledge of a convolutional neural network (CNN) model using random unlabeled data.
Another known attack mechanism is a Trojan attack described  in~\cite{liu2017trojaning} or a backdoor attack~\cite{gu2017badnets}. Such attacks are very dangerous, as they might cause various severe consequences, including endangering human lives, e.g., by disrupting the actions of a neural network-based autonomous vehicle. The idea is to spread and deploy infected models, which will act as expected for almost all regular inputs, except for a specific engineered input, i.e., a Trojan trigger, in which case the model would behave in a predefined manner that could become very dangerous in some cases. Consider, for example, an infected deep neural network (DNN) model of an autonomous vehicle, for which a specific given input will predict making a hard left turn. If such a model is deployed and triggered in the middle of a highway, the results could be devastating.

Using our composite method, even if our proposed student model learns from an infected mentor, it will not catch the dangerous triggers, and in fact, will act normally despite the engineered Trojan keys. The reason lies within our training method, as we randomly compose training examples based on the mentor's  prediction. In other words, the odds that a specific engineered key will be sent to the mentor and trigger a backdoor are negligible, similarly to the way training based on a mentor containing watermarks is done.

We present some interesting neural network attacks and show that our composite method is superior to these attacks and is also robust against infected models.

\subsection{Defense Mechanisms}
Besides watermarking, which is the main method of defending a model (or of enabling at least a stolen model to be exposed), there are some other available interesting possibilities.

In~\cite{2018ModelStealingAttack}, a method which adds a small controllable perturbation maximizing the loss of the stolen model while preserving the accuracy is suggested. For some attacking methods this trick can be effective and significantly slow down an attacker, if not prevent it completely. This method has no effect on our composite method, which preserves the accuracy, i.e., for each sample $x$, if for a specific index $i$ the softmax layer predicts $F(x)[i]$ as the maximum value, now the output of our network for that index would be $\hat{F}(x)[i] = F(x)[i] + \psi$, where $\psi$ is an intended perturbation, and where ${\rm argmax}(F(x)) = {\rm argmax}(\hat{F}(x)) = i$ still holds. This is the important element of our composite method, which solely relies on the model's binary labels and is not affected by this modification.

Most defense mechanisms are based mainly on manipulating the returned softmax confidence level, shuffling all of the label probabilities except for the maximal one, or returning a label without its confidence level. The base line is that all of these methods have to return the minimal information of what the predicted label is. Indeed, this is all that is required by the composite method, so our algorithm is unaffected by such defense mechanisms.

\section{Proposed Method}
In this section we present our novel composite method which can be used to attack and extract the knowledge of a black box model even if it completely conceals its softmax output.

For mimicking a mentor we assume no knowledge of the model's training data and no access to it (i.e., we make no use of any training data used to train the original model). Thus, the task at hand is very similar to real life scenarios, where there are plenty of available trained models (as services or products), without any knowledge of how they were trained and of the training data used in the process. Additionally, we assume no knowledge of the model's network architecture or weights, i.e., we regard it as an opaque black box. The only information about the model (which we would like to mimic) is its input size and the number of output classes (i.e., output size). For example, we may assume that only the input image size and the number of possible traffic signs are known, for a traffic sign classifier.

As previously indicated, another crucial assumption is that the black box model we aim at attacking does not reveal its confidence levels. Namely, the model's output is merely the predicted label, rather than the softmax values, e.g., in case of an input image of a traffic sign, whether the model is 99\% confident or only 51\% confident that the image is a stop sign, in both cases it will output ``stop sign'', without further information. We assume the model hides the confidence values as a safety mechanism against mimicking attacks by adversaries who are trying to acquire and copy the model's IP. Note that outputting merely the predicted class is the extreme protection possible for a model providing an API-based prediction, as it is the minimum amount of information the model must provide.

Our novel method for successfully mimicking a mentor that does not provide its softmax values, makes use of what we refer to as composite samples. By combining two different samples into a single sample (see details below) we effectively tap into the hidden knowledge of the mentor. (In the next section, we provide experimental results, comparing the performance of our method and that of standard mimicking using both softmax and non-softmax outputs.) 

For the rest of the discussion we refer to the black box model (we would like to mimic) and our developed model (for mimicking it) as a mentor model and a student model, respectively.

\subsection{Datasets for Mentor and Student}
\subsubsection{Dataset for Mentor Training}
CIFAR-10 \cite{cifar10} is an established dataset used for object recognition. It consists of 60,000 ($32 \times 32$) RGB images from 10 classes, with 6,000 images per class. There are 50,000 training images and 10,000 test images in the official data. The mentor is a pre-trained model on the CIFAR-10 dataset. We use the test set (from this dataset) to measure the success rate of our mentor and student models. Note that the training set of the CIFAR-10 dataset is never used in the training process by the student (to conform to our assumption that the student has no access to the data used by the mentor for training), and the test subset, as mentioned above, is used for validation only (without training).

\subsubsection{Dataset for Mimicking Process}
ImageNet~\cite{imagenet2009} is a dataset containing complex, real-world size images. In particular, ImageNet\_ILSVRC2012 contains more than 1.2 million ($256 \times 256$) RGB images from 1000 categories. We use this dataset (without the labels, i.e., an unlabeled dataset) for the mimicking process. Each image is down-sampled ($32 \times 32$) and fed into the mentor model, and the prediction of the mentor model is recorded (for later mimicking by the student). Note that any large unlabeled image dataset could be used instead, and we used this common large dataset for convenience only.

\subsection{Composite Data Generation}
Our goal is to create a diverse dataset that will allow to observe the predictions of the mentor on many possible inputs. By doing so, we would gain insights on the way the mentor behaves for different samples. That is, the more adequate the input space sample is, the better the performance of the mimicking process becomes.

The entire available unlabeled data, which is the down-sampled ImageNet, is contained in an array $dataArr$. For each training example to be generated,we choose randomly two indexes $i_1, i_2$, such that, $0 <= i_1, i_2 < N$, where $N$ is equal to the number of samples we create and use for training the student model. In our composite method we choose $N=1,000,000$, so the amount of generated training samples created in each epoch is $1,000,000$. Next, we randomly choose a ratio $p$. Once we have $i_1, i_2$ and $p$, we generate a composite sample, created by combining two existing images in the dataset. The ratio $p$ determines the relative influence of the two random images on the generated sample:
$$x\_gen = p*dataArr[i_1]+(1-p)*dataArr[i_2],$$
where the label of $x\_gen$ is a ``one-hot'' vector, i.e., the index containing the '1' (corresponding to the maximal softmax value) represents the label predicted by the mentor. The dataset is generated for every epoch; hence, our composite dataset changes continuously and it is dynamic. We gain the predictions of a mentor model on new images during the entire training process (with less overfitting).

Note that even though in our data-generating mechanism we create a composite of two random images (with a random mixture between them), it is possible to create composite images of $N$ images where $N>2$, as well.

\begin{figure}
\centering
\begin{subfigure}{.45\linewidth}
\centering
\includegraphics[width=.9\linewidth]{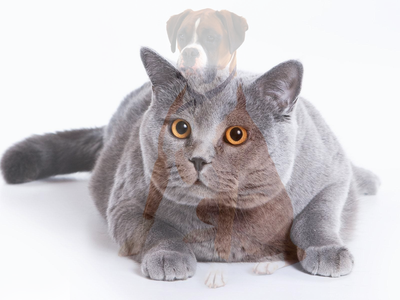}
\caption{75\% cat 25\% dog}
\label{sfig:incorrect_a}
\end{subfigure}
\begin{subfigure}{.45\linewidth}
\centering
\includegraphics[width=.9\linewidth]{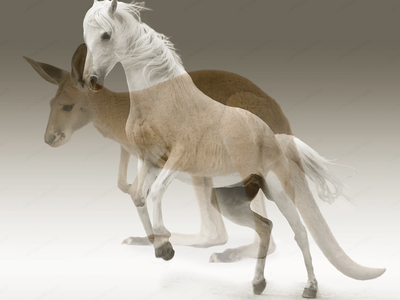}
\caption{70\% horse 30\% kangaroo}
\label{sfig:incorrect_b}
\end{subfigure}\par\medskip
\begin{subfigure}{.45\linewidth}
\centering
\includegraphics[width=.9\linewidth]{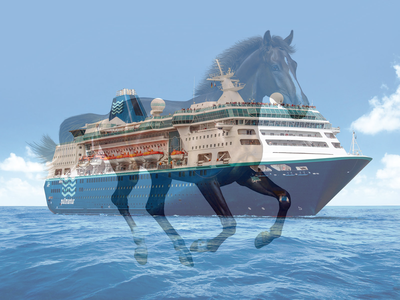}
\caption{30\% horse 70\% ship}
\label{sfig:incorrect_c}
\end{subfigure} %
\begin{subfigure}{.45\linewidth}
\centering
\includegraphics[width=.9\linewidth]{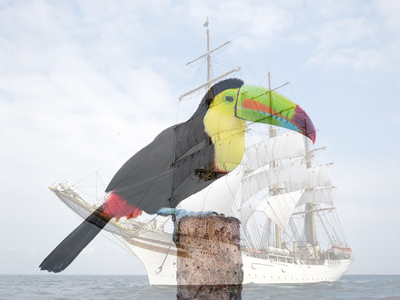}
\caption{40\% ship 60\% parrot}
\label{sfig:incorrect_d}
\end{subfigure}\par\medskip
\begin{subfigure}{.45\linewidth}
\centering
\includegraphics[width=.9\linewidth]{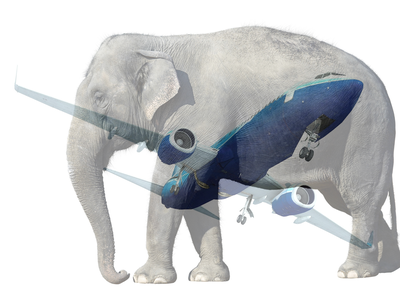}
\caption{50\% elephant 50\% airplane}
\label{sfig:incorrect_e}
\end{subfigure}%
\begin{subfigure}{.45\linewidth}
\centering
\includegraphics[width=.9\linewidth]{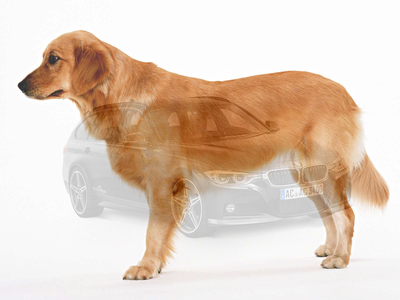}
\caption{80\% dog 20\% car}
\label{sfig:incorrect_f}
\end{subfigure}
\begin{subfigure}{.45\linewidth}
\centering
\includegraphics[width=.9\linewidth]{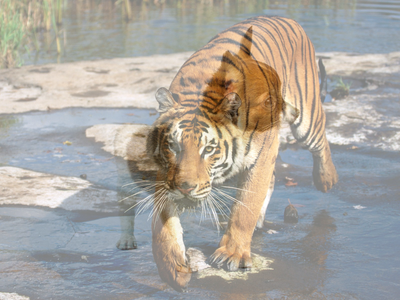}
\caption{50\% tiger 50\% dog}
\label{sfig:incorrect_e}
\end{subfigure}%
\begin{subfigure}{.45\linewidth}
\centering
\includegraphics[width=.9\linewidth]{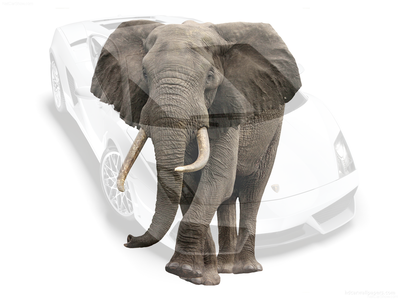}
\caption{20\% car 80\% elephant}
\label{sfig:incorrect_f}
\end{subfigure}

\caption{Illustration of images created using our composite data-generation method. The images and their relative mixture are random. Using this method during each epoch we create an entirely new dataset, with random data not seen before by the model.}
\label{fig:composite_dat}
\end{figure}

Algorithm~\ref{alg:data_Gen} provides the complete composite data-generation method, which is run at the beginning of each epoch. Figure~\ref{fig:composite_dat} is an illustration of composite data samples created by Algorithm~\ref{alg:data_Gen}.

\newcommand{\vars}{\texttt}
\newcommand{\func}{\textrm}
\let\oldReturn\Return
\renewcommand{\Return}{\State\oldReturn}
\begin{algorithm}
\caption{Composite Data Generation}
\label{alg:data_Gen}
\begin{algorithmic}[1]
\State \textbf{\func{Input:}}
\State $mentor$ - \func{the mentor model}
\State $dataArr$ - \func{all available data array}
\State $N$ - \func{number of samples to generate}
\State \textbf{\func{Output:}}
\State $X$ - \func{generated examples}
\State $Y$ - \func{corresponing labels}
\Function{GENERATE\_DATA}{mentor, dataArr, N}
\State $X,Y = [],[]$
\For{i=1 to N}
\State \vars{$i_1$ = \func{math.random}$($\func{len}\vars{$(dataArr))$}}
\State \vars{$i_2$ = \func{math.random}$($\func{len}\vars{$(dataArr))$}}
\State \vars{$p$ = \func{math.random}\vars{(100)}/100}
\State \vars{$x\_gen$ = $p*dataArr[i_1]+(1-p)*dataArr[i_2]$ }
\State \vars{$X.$}\func{append}\vars{$(x\_gen)$}
\State \vars{$Y.$}\func{append}$($\func{argmax}$($\vars{$mentor$}\func{.predict}\vars{$(x\_gen)$}$))$
\EndFor
\Return $X,Y$
\EndFunction
\end{algorithmic}
\end{algorithm}

\subsection{Student Model Architecture}
The mentor neural network (which we intend to mimic) is an already trained model that reaches 90.48\% test accuracy on the CIFAR-10 test set. 

Our goal in choosing an architecture for the student is to be generic, such that it would perform well, regardless of the mentor we try to mimic. Thus, with small adaptations to the input and output size, we created a modification of the VGG-16 architecture~\cite{simonyan2014very} for the student model. In our model we use two dense layers of size 512 each, and another dense layer of size 10 for the softmax output (while in the original VGG-16 architecture, there are two dense layers of size 4096 and another dense layer of size 1000 for the softmax layer). Table~\ref{table:model_architecture} presents the architecture of our student model.

\begin{table}[h]
\begin{center}
\def\arraystretch{1.3}%
\begin{tabular}{c}
    \thickhline
	\setrow{\bfseries}Modified VGG-16 Model Architecture for Student Network\\
    \hline
    3x3 Convolution 64 \\
    3x3 Convolution 64 \\
    \hline
    Max pooling\\
    \hline
    3x3 Convolution 128 \\
    3x3 Convolution 128 \\
    \hline
    Max pooling\\
    \hline
    3x3 Convolution 256 \\
    3x3 Convolution 256 \\
    3x3 Convolution 256 \\
    \hline
    Max pooling\\
    \hline
    3x3 Convolution 512 \\
    3x3 Convolution 512 \\
    3x3 Convolution 512 \\
    \hline
    Max pooling\\
    \hline
    3x3 Convolution 512 \\
    3x3 Convolution 512 \\
    3x3 Convolution 512 \\
    \hline
    Max pooling\\
    \hline
    Dense 512\\
    Dense 512\\
    \hline
    Softmax 10\\
    \thickhline
\end{tabular}
\caption{The architecture used in the composite training experiment for the student model. This architecture is a modification of the VGG-16 architecture which has proven to be very successful and robust. By performing only small modifications over the input and output layers, we can adapt this architecture for a student model intended to mimic a different mento model.}
\label{table:model_architecture}
\end{center}
\end{table}

\subsection{Mimicking Process}
Using the above described composite data generation, a new composite dataset is generated for every epoch during the mimicking process. We train on this dataset using the stochastic gradient descent (SGD) algorithm. Table~\ref{table:composite_params} describes the parameters used for training the student model.

Our student model does not use any dropout or regularization methods. Such regularization methods are not necessary since our model does not reach overfitting as a result of the dynamic dataset (a new composite dataset generated at each epoch). To evaluate the final performance of the student model, we test it on a dedicated test set that was used to evaluate also the mentor model. (Neither the student nor the mentor were trained on images belonging to the test set.)

We also used learning rate decay, starting from 0.001 and multiplied by 0.9 every 10 epochs, as we found it essential in order to reach high accuracy rates. In Section IV we provide a detailed description of our experimental results.

\begin{table}[h]
\begin{center}
\def\arraystretch{1.3}%
\begin{tabular}{lc}
    \thickhline
	\setrow{\bfseries}Parameters &\setrow{\bfseries} Values\\
    \thickhline
    Learning rate & 0.001\\
    Activation function & ReLU\\
    Batch size & 128\\
    Dropout rate & -\\
    $L_2$ regularization & -\\
    SGD momentum & 0.9\\
    Data augmentation & -\\
    \thickhline
\end{tabular}
\caption{Parameters used for training in the composite experiment.}
\label{table:composite_params}
\end{center}
\end{table}

\subsection{Data Augmentation}
Data augmentation is a useful technique frequently used in the training process of deep neural networks. It is mostly used to synthetically enlarge a limited size dataset, in an attempt to generalize and enhance the robustness of a model under training, and to reduce overfitting.

The basic idea is very simple. The model is not trained on the same training samples at each epoch. Instead, during each epoch small random visual modifications are made to the dataset images. This is done in order to allow the model to be trained during each epoch on a slightly different dataset, using the same labels for the training. Examples of simple data augmentation operations include small vertical and horizontal shifts of the image, a slight rotation of the image (usually by $\theta$ for $0^\circ < \theta <= 15^\circ$), etc.

This technique is used for our student models, which are trained on the same dataset during each epoch. However, for the composite model experiment, we found it to have no effect on the performance. Our composite data-generation method ensures virtually a continuous set of infinitely many new samples never seen before; thus, data augmentation is not necessary here at all.

Our end goal is to represent a nonlinear function, which takes an $n$-dimensional input and transforms it to an $m$-dimensional output, e.g., a function that takes an image of size $256\times256$ of a road and returns one of $Y$ possible actions that an autonomous vehicle should take. Using data augmentation, we can train the model to better represent the required nonlinear function. For our composite method, though, this would be redundant, since the training process is always performed on different random inputs, which allows for estimating empirically the nonlinear function in a much better way, without using the original training dataset for training the model.

\section{Experimental Results}
\subsection{Experimental Results for Unprotected Mentor (with Softmax Output) and Standard Mimicking}

To obtain a baseline for comparison, we assume in this experiment that the mentor in question reveals its confidence levels by providing the values of its softmax output. (We refer to it as an ``unprotected mentor''.)

We use here the same modified VGG-16 architecture presented in Table \ref{table:model_architecture}.

We create, in this case, a new dataset for the student model only once and use it together with standard data augmentation. We feed each training sample from the down-sampled ImageNet into the mentor, and save the pairs of its input image and softmax label distribution (i.e., its softmax layer output). The total size of this dataset is over 1.2 million samples (the size of the ImageNet\_ILSVRC2012 dataset). Once the dataset is created, we train the student using regular supervised training with SGD. In this experiment, since overfitting would occur without regularization, we use dropout to improve generalization. The parameters used for training this model are presented in Table~\ref{table:available_distribution_params}.

Using these parameters, we obtained a maximum test accuracy of 89.1\% for the CIFAR-10 test set, namely, 1.38\% less than the mentor's 90.48\% success rate. (Note that the student was never trained on the CIFAR-10 dataset, and instead, after completing the mimicking process using the separate unrelated dataset, its performance was only tested on the CIFAR-10 test set.)

\begin{table}[h]
\begin{center}
\def\arraystretch{1.3}%
\begin{tabular}{lc}
    \thickhline
	\setrow{\bfseries}Parameters &\setrow{\bfseries} Values\\
    \thickhline
    Learning rate & 0.001\\
    Activation function & ReLU\\
    Batch size & 128\\
    Dropout rate & 0.2\\
    $L_2$ regularization & 0.0005\\
    SGD momentum & 0.9\\
    Data augmentation & Used\\
    \thickhline
\end{tabular}
\caption{Parameters used for the training process using standard (non-composite) mimicking.}
\label{table:available_distribution_params}
\end{center}
\end{table}

\subsection{Experimental Results for Protected Mentor (without Softmax Output) and Standard Mimicking}

In this experiment we assume that the mentor reveals the predicted label with no information about the certainty level (i.e., it is considered a ``protected mentor''). This is a real-life scenario, in which an API-based service is queried by uploading inputs, and only the predicted output class (without softmax values) is returned. 

By sending only the correct labels, the models are more protected, in the sense that they reveal less information to a potential attacker. For this reason, this method has become a common defense mechanism for protecting intellectual property when neural networks are deployed in real-world scenarios.

In this subsection we try a standard mimicking attack (without composite images). Here we execute exactly the same training process of the soft labels experiment (described in the previous subsection) with one important difference. In this case, the labels available for the student are merely one-hot labels provided by the mentor, and not the full softmax distribution of the mentor. For each training sample (from the down-sampled ImageNet dataset), we take the output distribution, find the index with the maximum value, and set it to '1' (while setting all the other indices to '0'). The student can observe only this final vector, with a single '1' for the correct class, and '0' for all other classes. This accurately simulates a process that can be applied on an API service.

The student only knows at this point the mentor's prediction, but not its level of certainty. We use the same parameters of  Table~\ref{table:available_distribution_params} for the mimicking process.

The success rate in this experiment is the lowest; the student reached only $\sim 87.5\%$ accuracy on the CIFAR-10 test set, i.e., substantially lower than that of the student which mimicked an unprotected mentor.

\subsection{Experimental Results for Protected Mentor (without Softmax Output) and Composite Data Mimicking}

In this experiment we assume again that our mentor reveals the predicted label with no information about the certainty level. However, instead of launching a standard attack on the mentor, we employ here our novel composite data generation as described in Algorithm~\ref{alg:data_Gen}, in order to generate new composite data samples at each epoch. In this case, the student only has access to the predicted labels (minimum output required from a protected mentor).

Unlike the previous two experiments using standard mimicking, we do not use here data augmentation or regularization, since virtually all of the data samples are always new, and are generated continuously.

\begin{figure}
\centering

\begin{subfigure}{.35\linewidth}
\centering
\includegraphics[width=.9\linewidth]{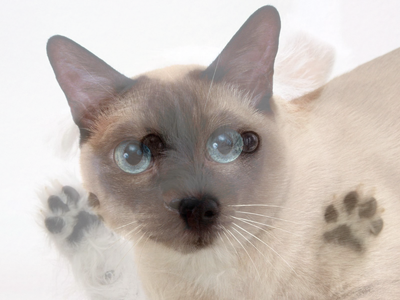}
\caption{50\% dog 50\% cat}
\label{sfig:incorrect_a}
\end{subfigure}
\begin{subfigure}{.635\linewidth}
\centering
\includegraphics[width=.9\linewidth]{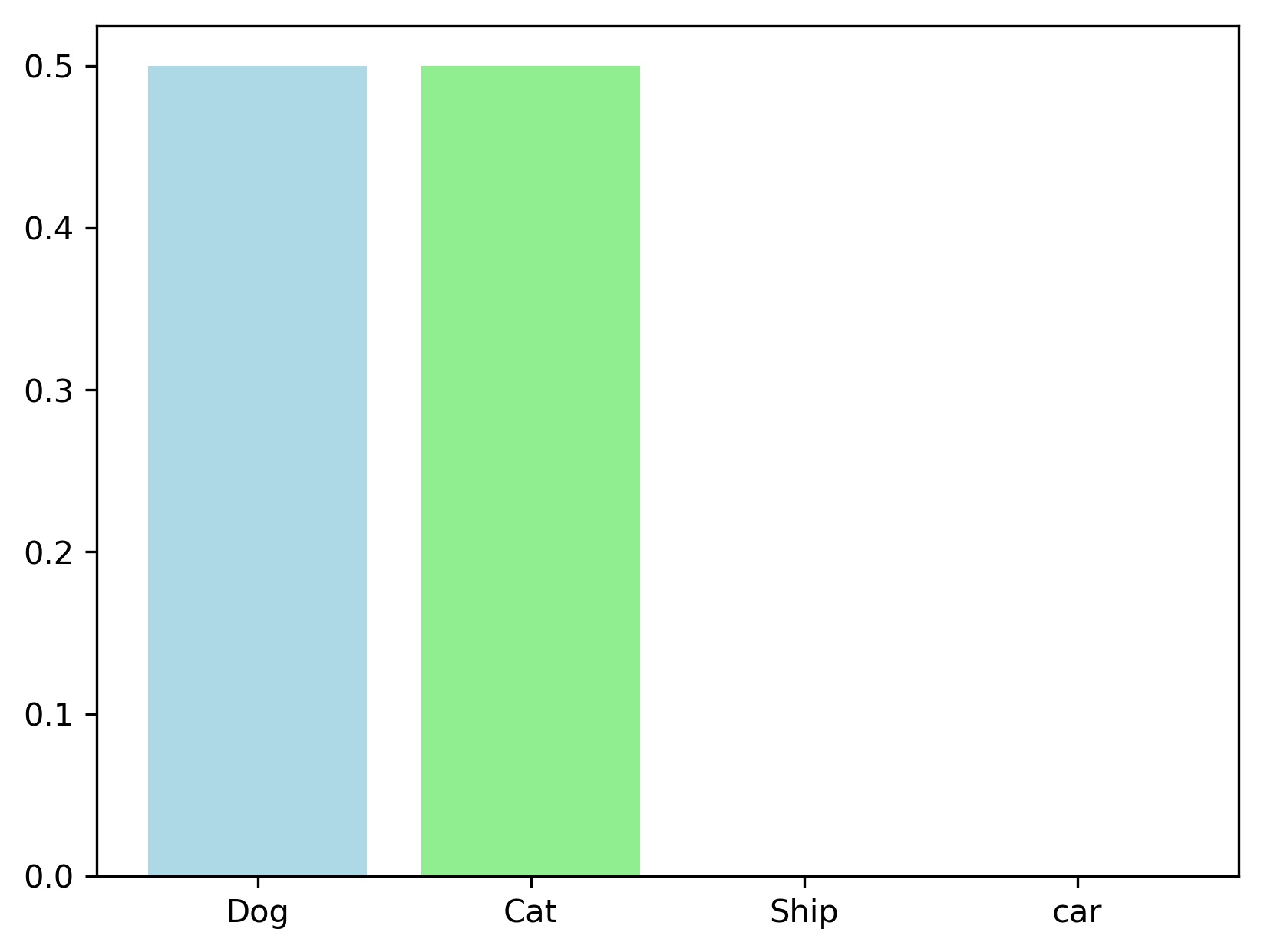}
\label{sfig:incorrect_b}
\end{subfigure}\par\medskip

\begin{subfigure}{.35\linewidth}
\centering
\includegraphics[width=.9\linewidth]{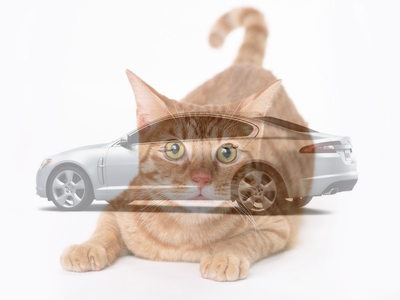}
\caption{60\% cat 40\% car}
\label{sfig:incorrect_a}
\end{subfigure}
\begin{subfigure}{.635\linewidth}
\centering
\includegraphics[width=.9\linewidth]{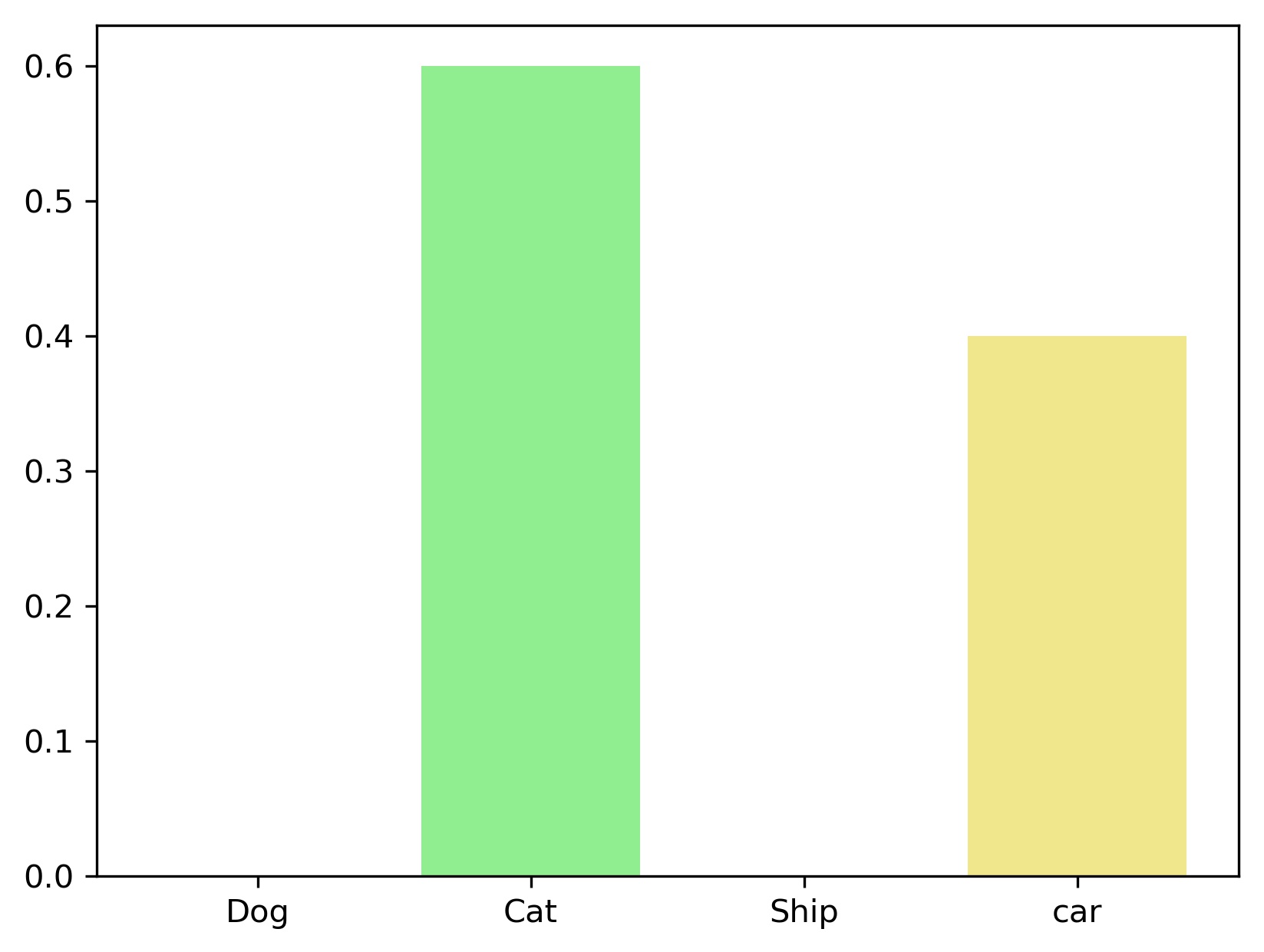}
\label{sfig:incorrect_b}
\end{subfigure}\par\medskip

\begin{subfigure}{.35\linewidth}
\centering
\includegraphics[width=.9\linewidth]{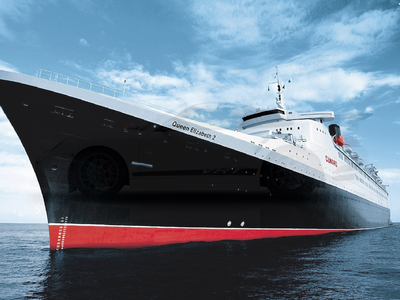}
\caption{90\% ship 10\% car}
\label{sfig:incorrect_a}
\end{subfigure}
\begin{subfigure}{.635\linewidth}
\centering
\includegraphics[width=.9\linewidth]{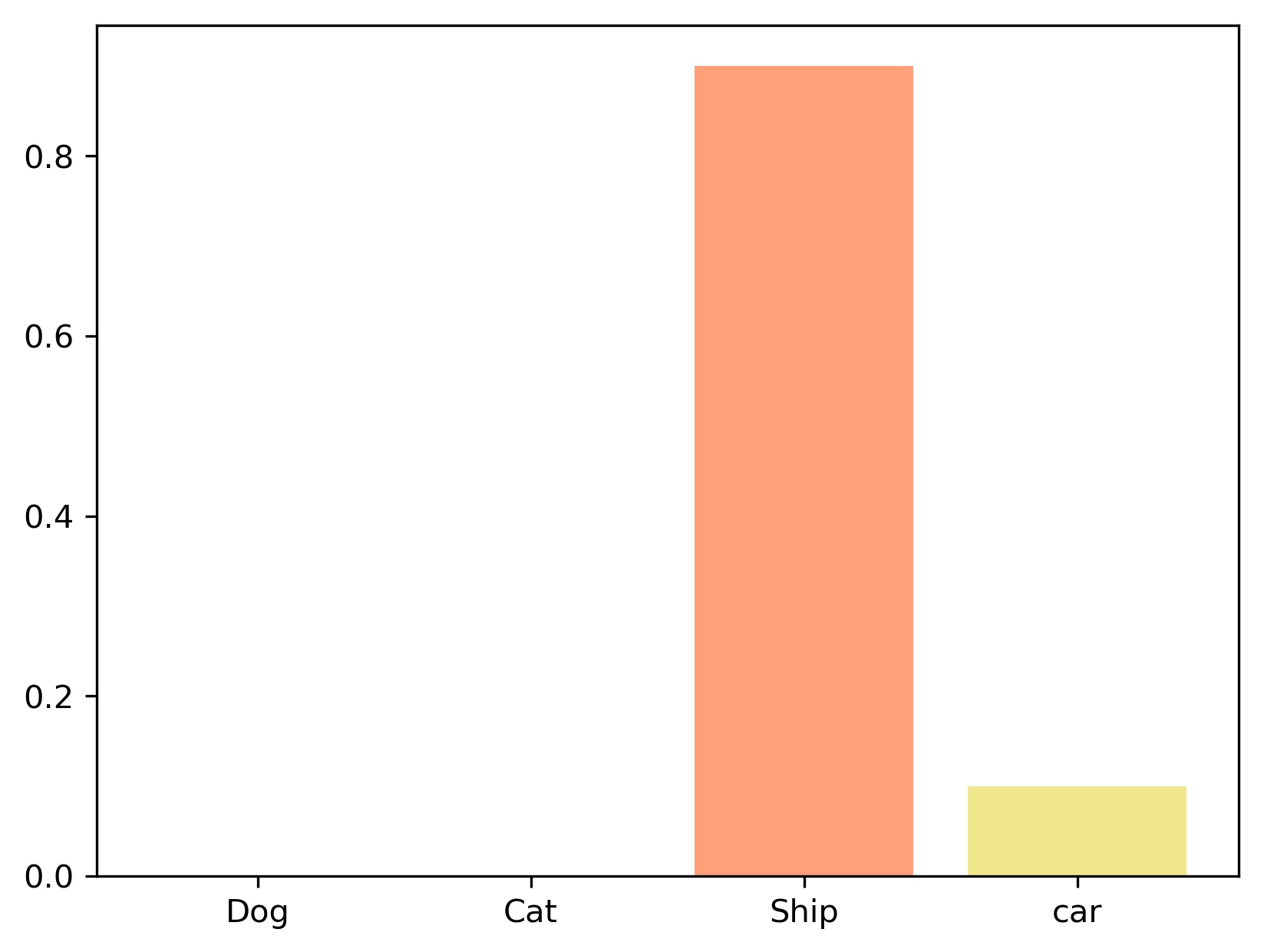}
\label{sfig:incorrect_b}
\end{subfigure}\par\medskip

\caption{Generated images and their corresponding expected softmax distribution, which reveals the model's certainty level for each example. In practice, the manner by which objects overlap and the degree of their overlap largely affect the certainty level.}
\label{fig:composite_bars}
\end{figure}

Figure~\ref{fig:composite_bars} illustrates the expected predictions from a well-trained model for certain combined input images. Empirically, this is not totally accurate, since the presentation and overlap of objects in an image also affect the output of the real model. However, despite this caveat, the experimental results presented below show that our method provides a good approximation.

Training with composite data, we obtained 89.59\% accuracy on the CIFAR-10 test set, which is only 0.89\% less than that of the mentor itself. (Again, note that the student is not trained on any of the CIFAR-10 images, and that the test set is used only for the final testing, after the mimicking process is completed.) This is the highest accuracy among all of the experiments conducted; surprisingly, it is even superior to the results of standard mimicking for an unprotected mentor (which does divulge its softmax output). 

\begin{figure}[h]
  \centering
  \hspace*{-0.2cm}\includegraphics[width=9cm,height=7cm]{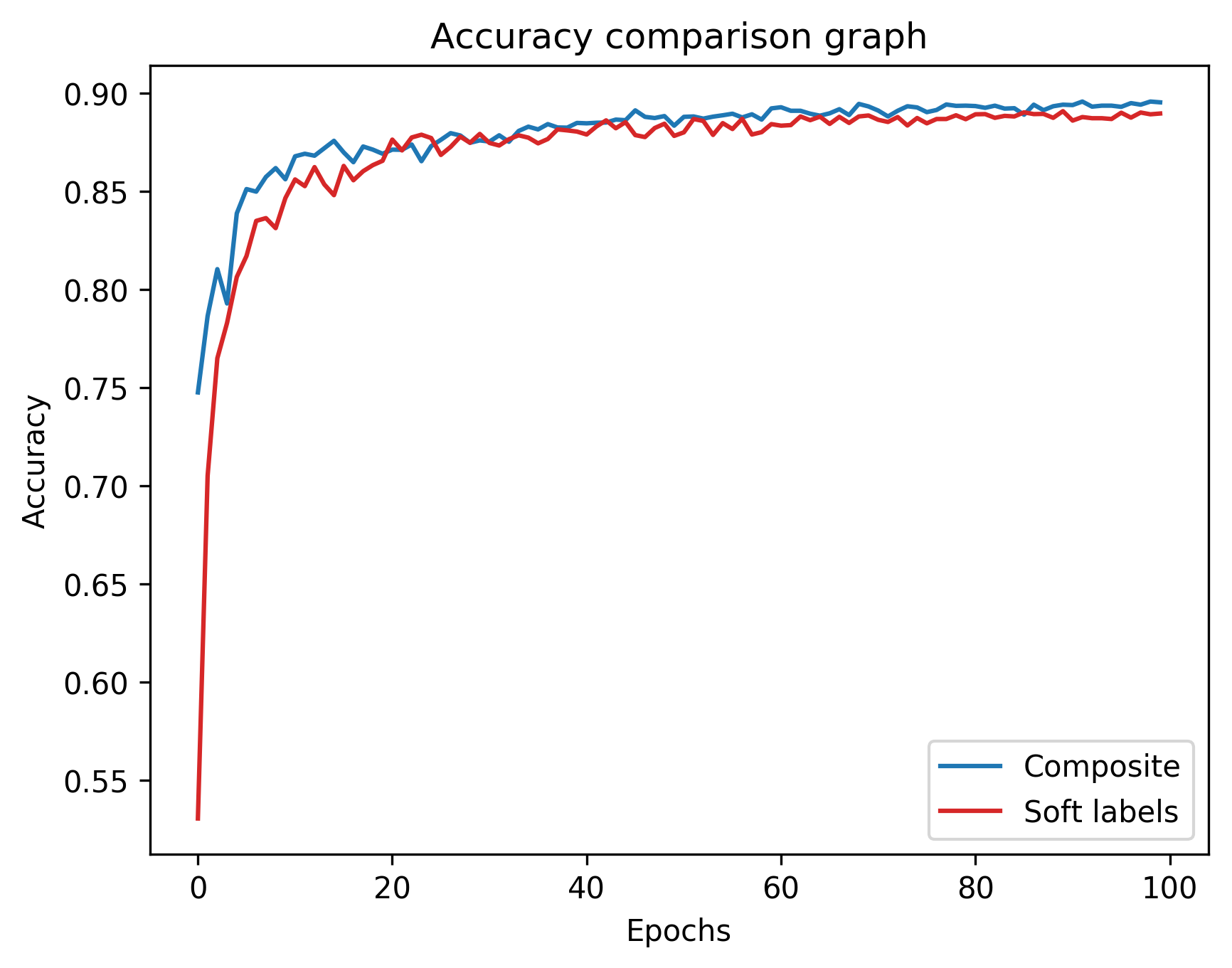}
  \caption{Student test accuracies for composite and soft-label experiments, training the student over 100 epochs. The student trained using the composite method is superior during almost the entire training process. The two experiments were selected for visual comparison as they reached the highest success rates for the test set.}
  \label{fig:accuracies_comparison}
\end{figure}

Figure~\ref{fig:accuracies_comparison} depicts the accuracy over time (i.e., epoch number) for the composite and soft-label experiments.
As can be seen, the success rate of the composite experiment is superior to that of the soft-label experiment during almost the entire training process. Even though the latter has access to valuable additional knowledge, our composite method performs consistently better without access to the mentor's softmax output.

A summary of the experimental results is presented in  Table~\ref{table:results_summary}, including relative accuracy to the mentor's accuracy rate. The results show that standard mimicking obtained $\sim98.5\%$ of the accuracy of an unprotected mentor, and only $\sim96.7\%$ of its accuracy when the mentor was protected. However, using the composite mimicking method, the student reached (over) 99\% of the accuracy of a fully protected mentor.

Thus, even when a mentor only reveals its predictions without their confidence levels, the model is not immune to mimicking and knowledge stealing. Our method is generic, and can be used on any model with only minor modifications on the input and output layers of the architecture.

\begin{table}[h]
\begin{center}
\def\arraystretch{1.5}%
\begin{tabular}{cccc}
    \thickhline
	\setrow{\bfseries}Method &\setrow{\bfseries} Mentor Status &\setrow{\bfseries} Test Accuracy &\setrow{\bfseries} Relative Accuracy\\
    \thickhline
    Standard & Unprotected & 89.10\% & 98.47\%\\
    Standard & Protected & 87.46\% & 96.66\%\\
    \setrow{\bfseries}Composite &\setrow{\bfseries} Protected &\setrow{\bfseries} 89.59\% &\setrow{\bfseries} 99.01\%\\
    \thickhline
\end{tabular}
\caption{Summary of the experiments. The table provides the CIFAR-10 test accuracy of three student models in absolute terms and in comparison to the 90.48\% test accuracy achieved by the mentor itself. The three mimicking methods use standard mimicking for unprotected and protected mentors, as well as composite mimicking for a protected mentor, which provides the best results.}
\label{table:results_summary}
\end{center}
\end{table}

\section{Conclusion}
In view of the tremendous DNN-based advancements in various hard, non-trivial problem domains in recent years, the issue of protecting complex DNN models has gained considerable interest. The computational power and significant effort required by a training process makes a well-trained network very valuable. Thus, much research has been devoted to attacks on DNNs and to their defense, where the most common defense mechanism is to conceal the model's certainty levels and output merely a predicted label.

In this paper, we have presented a novel composite image attack method for extracting the knowledge of a DNN model, which is not affected by the above ``label only'' defense mechanism. Specifically, our composite method assumes only that this mechanism is activated and relies solely on the label prediction returned from a black box model. We assume no knowledge about this model's training process and its original training data. In contrast to other methods suggested for stealing or mimicking a trained model, our method does not rely on the softmax distribution supplied by a trained model with a certainty level across all categories. Therefore, methods adding a controlled perturbation to the returned softmax distribution in order to protect a given model are essentially ineffective against our method.

By employing our method, a user can attack a model and reach a very close success rate compared to it, while relying only on the minimal information that has to be given by the model, namely its label prediction for a given input. Our method demonstrates that the current available defense mechanisms for deep neural networks are not sufficient. Using the composite attack method, countless neural networks-based services can be attacked and copied into a rival model which can then be deployed and affect the product's market share. The rival deployed model will be undetectable and carry no mark proofs that it is stolen, as explained in Section II(A).


\bibliographystyle{IEEEtran}
\bibliography{IEEEabrv,stealing-knowledge}
\end{document}